\documentclass[pmlr,twocolumn,10pt]{jmlr}

\mlhtrack{proceedings}

\usepackage{booktabs}
\usepackage{multirow}
\usepackage{graphicx}
\usepackage{heatcell}
\definecolor{styleorange}{RGB}{255,140,0}
\definecolor{darkgreen}{RGB}{0,100,0}
\definecolor{altblue}{RGB}{20,60,160}

\usepackage[breakable,skins]{tcolorbox}

\newheatcol{F}{min=25, max=90, color=cyan}
\newheatcol{M}{min=0.1, max=2.7, color=red} % MAE, lower is better
\newheatcol{R}{min=0.3, max=0.9, color=green}          % correlation
\newheatcol{A}{min=55, max=90, color=cyan}            % direction accuracy (%)
\newheatcol{Z}{min=1, max=4, color=styleorange}       % style score (1--5 rubric); avoid S which collides with siunitx
\usepackage{siunitx}
\usepackage{enumitem}
\usepackage[switch]{lineno}
\usepackage[all]{hypcap}

\title[Decomposing Error and Style]{Decomposing Error and Style in Automated Clinical Coding}

\author{%
 \Name{Han-Chin Shing$^{*}$} \Email{hanchins@amazon.com}
 \AND
 \Name{Jack Moriarty} \Email{}
\AND
 \Name{Ryan Ware} \Email{}
 \AND
 \Name{Afton Marchbanks} \Email{}
 \AND
 \Name{Carlyn Canvasser} \Email{}
 \AND
 \Name{Stefanie Higgins} \Email{}
 \AND
 \Name{Harsh Gupta} \Email{}
 \AND
 \Name{Fang Wang} \Email{}
 \AND
 \Name{Joseph Paul Cohen$^{*}$} \Email{joseph@josephpcohen.com}\\
 \addr Amazon\\
 \addr $^{*}$Equal contribution.
}

\begin{document}

\maketitle

\begin{abstract}
In automated clinical coding, where the label space spans tens of thousands of diagnosis and procedure codes, models are currently evaluated against a single gold annotation, treating any deviation as error. But we find when two teams code the same 110 ACI-Bench encounters, they agree on only $73\%$ of codes (Jaccard similarity) for the same note; even after an independent clinical audit removes erroneous codes, agreement rises only to $77\%$. Is that gap error or something systematic?
We model the systematic component as \emph{coding style} $\psi$, a coder- or site-specific policy over what to code and how much to document, and recast coding as $p(\mathrm{code}\mid\mathrm{note},\psi)$, estimating $\psi$ with a 10-dimension rubric. If style were noise, conditioning on it would do nothing. Instead, across five datasets a model conditioned with a data-matching style raises ICD F1 by up to 26 points and an extreme mismatched one lowers it by up to 21. Four prompt based coding methods spanning 39--49 F1 converge to 52--56 once style is supplied (All $p<0.05$). Much of what single-gold evaluation charges to model error is recoverable, unmodeled style.

\end{abstract}

\begin{keywords}
Clinical coding, ICD-10, large language models, style adaptation,
label variation, evaluation
\end{keywords}
\begin{figure*}[t]
\centering 
\includegraphics[width=1\textwidth]{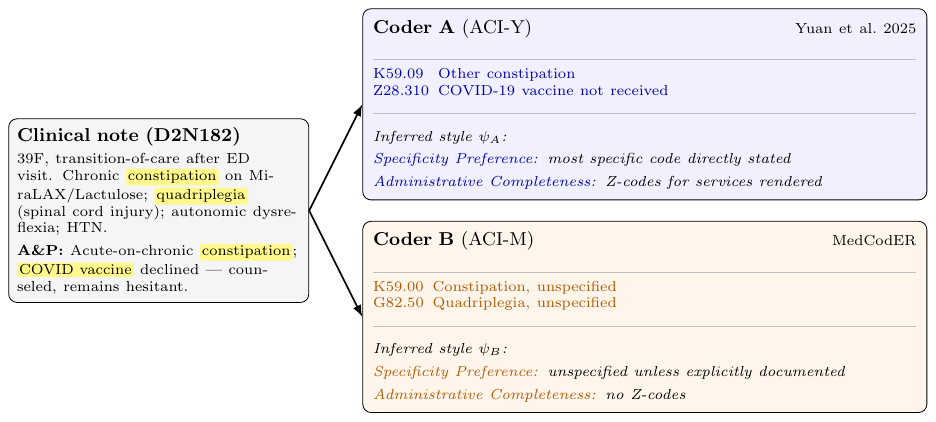}
\caption{Same note (summarized), two coders: disjoint code sets, all four codes judged defensible by a professional clinical-coding auditor. The gap is style, not clinical disagreement. Two coders coded the same chief complaint with differing specificity (K59.09 vs. K59.00) and administrative completeness (Z28.310 vs. the quadriplegia comorbidity G82.50).}
\label{fig:teaser}
\end{figure*}

\section{Introduction}
\label{sec:intro}
Automated clinical coding, the task of translating clinical notes into alphanumeric codes for billing and reimbursement purposes, is usually framed as an extreme multi-label classification problem measured with F1 against a single gold annotation. This hides the assumption that every competent coder, given the same note and same guidelines, would produce the same codes. In practice they do not: even expert coders working from identical notes routinely disagree \citep{crammer2007automatic, minarro2018qualitative}. This is unsurprising in a label space of roughly 72{,}000 billable ICD-10-CM codes (diagnoses) and 10{,}000 CPT codes (procedures) \citep{cms2024icd10cm}, where coders rely on learned shortcuts and local convention.

How much do they disagree, and why? On 110 public ACI-Bench encounters, two independent annotation teams (ACI-Y, ACI-M) agree on only $73\%$ of their codes (Jaccard similarity) despite coding identical notes, and an independent clinical audit that removes not-defensible codes only lifts this to $77\%$ (Appendix~\ref{app:audit}). When the note is held fixed, one hypothesis, dominant in the literature, treats this gap as error and inter-annotator variation, an irreducible cost of the expertise a vast code space demands. We examine a second, largely overlooked hypothesis, echoing a broader shift toward treating human label variation as signal rather than noise \mbox{\citep{plank2022problem, weber2024varierr}}: that part of the disagreement is systematic. We call this systematic component \emph{coding style}, a coder- or site-specific policy over what to code and how much documentation to require; the name is a matter of convenience, and the same concept could be called a coder's policy, philosophy, or conventions. The two hypotheses are not exclusive: real disagreement is some mixture of error and style. Figure~\ref{fig:teaser} shows one case in which the same note receives disjoint code sets from the two teams, yet a professional clinical-coding auditor judged every code defensible (Appendix~\ref{app:qualitative}).

The distinction matters because the two are measured and remedied differently. If disagreement were purely error, better coders or trained models would converge toward the single gold set. Instead, prior systems have plateaued and transfer poorly: supervised label-attention encoders \citep{mullenbach2018explainable}, direct LLM prompting \citep{boyle2023automated}, and retrieval-plus-verification pipelines \citep{yuan2025verify} are all trained and scored against one gold set per note, and their accuracy degrades across MIMIC versions \citep{edin2023mimic}, across hospitals \citep{ponthongmak2023external}, and across admission years \citep{blanco2020extreme}. The common response is to fine-tune a separate model per site, which is expensive and treats a site's conventions as something to learn into the weights of the model rather than describe. 

Prior work in health economics has examined how payment models relate to coding behavior. Fee-for-service payment does not directly tie most provider payments to reported diagnoses, whereas risk-adjusted programs such as Medicare Advantage do. \citet{geruso2020upcoding} report that enrollees in private Medicare plans generate 6--16\% higher diagnosis-based risk scores than under fee-for-service, and \citet{kronick2014measuring} report that coding intensity varies widely by contract. We cite these studies only to note that coding behavior has been observed to vary across settings. We do not attempt to explain what drives style in our data; instead, we ask a narrower and testable question: whether modeling style, whatever its origin, is useful for coding.

If part of the disagreement is systematic, it should be measurable and controllable, and single-gold F1 should be understood as scoring style alignment as much as coding ability. We make this concrete by casting coding as $p(\mathrm{code}\mid\mathrm{note},\psi)$, where guidelines constrain which codes are admissible and style $\psi$ selects among them. We make $\psi$ measurable with a 10-dimension rubric, estimate it for each data source with a style profiler, and condition a prompt-based coder on the estimated profile at inference time, with no fine-tuning. Four coding methods that range from $39$--$49$ F1 without style converge to $52$--$56$ F1 once a matched style is supplied, indicating that much of the gap previously attributed to model error is unmodeled style.

We do not claim that ten dimensions capture style completely; they are deliberately coarse, and ACI-Y and ACI-M yield nearly identical averaged profiles despite disagreeing on a quarter of their codes. Because even this simplified rubric accounts for a substantial share of the gap and improves every method we test, we present our results as a lower bound on what modeling coding style can recover, with the remainder marking where genuine error begins.

% {\color{altblue}%
We make five contributions.
\textbf{(C1) Framing.} We recast clinical coding as $p(\mathrm{code}\mid\mathrm{note},\psi)$, separating the gap against a single gold set into genuine error and \emph{coding style} $\psi$, where guidelines fix which codes are admissible and $\psi$ selects among them.
\textbf{(C2) Rubric.} We make $\psi$ measurable with a 10-dimension rubric (6 ICD-scope, 4 CPT-evidence; Section~\ref{sec:rubric}) estimated from (note, code-set) pairs by a style profiler.
\textbf{(C3) Separability.} Per-encounter style vectors cluster by source (Figure~\ref{fig:tsne}), indicating that style is a property of the data source rather than per-note noise.
\textbf{(C4) Steerability.} Conditioning a coder on a style profile at inference time, without training, changes ICD F1 by up to $+26$ for a matched profile and up to $-21$ for a mismatched one.
\textbf{(C5) Generality.} The effect holds across coding methods: adding a style block improves two published pipelines, Two-Step \citep{boyle2023automated} and Yuan-Verify \citep{yuan2025verify}, by $18.1$ and $16.0$ points, and four coders spanning $39$--$49$ F1 converge to $52$--$56$ F1 once style is supplied.
% }

\section{Framing}
\label{sec:approach}

We treat clinical coding as
\begin{equation}
\label{eq:coder}
p(\mathrm{code} \mid \mathrm{note}, \psi),
\end{equation}
where $\psi$ defines what to include and how much documentation is required. Coding guidelines constrain the support of $p$ while $\psi$ shapes the choice within that support. Under this view, a ``wrong'' prediction against a single gold set may in fact be a valid draw from a different $\psi$. A \emph{style profiler} runs the other direction, recovering the style that a set of codes reflects:
\begin{equation}
\label{eq:profiler}
p(\psi' \mid \mathrm{note}, \mathrm{code}),
\end{equation}
where $\psi'$ is the style inferred from an observed (note, code) pair. The true style behind a coder's decisions is high-dimensional and only partly observable. Rather than model it exactly, we approximate $\psi$ with a compact, human-interpretable rubric, coarse by design, but enough to describe and transfer the dominant regularities. 

Our Approach:

\begin{enumerate}[noitemsep,topsep=2pt] 
\item \textbf{Define} $\psi$ as a 10-dimension rubric (6 ICD-scope dims, 4 CPT-evidence dims), designed to describe coding style.
\item \textbf{Estimate} $\psi'$ per data source with a \emph{style profiler}: an LLM scores 50--150 annotated encounters and the exercised (non-N/A) scores are averaged into an integer profile.
\item \textbf{Condition} models on $\psi'$ at inference time by prepending a natural-language \texttt{<coding\_style>} block derived from $\psi'$ (\S\ref{app:prompt}); no fine-tuning, no per-site model.
\end{enumerate}

\section{Rubric}
\label{sec:rubric}

The rubric has 10 dimensions (6 ICD, 4 CPT), shown in Table~\ref{tab:rubric}. Each is scored 1--5 by a style profiler. \textbf{ICD and CPT dimensions (S, C)} are scored from the note and the associated codes (typically ground truth); these estimated styles can be applied at inference time.

During style determination the profiler can say N/A for any dimension not exercised by the encounter; only exercised scores enter the averaged profile. At inference time the full averaged profile is always presented with an ``if applicable'' framing, so nothing about the encounter type leaks through which dimensions appear.

Two caveats. First, the 1--5 Likert scale is convenient to average and plot, but we do not expect the coder LLM to follow it linearly; each score maps to an explicit written behavior description (\S\ref{app:prompt}), and discrete written categories per dimension (e.g., ``codes from the imported problem list'' vs.\ ``codes only newly addressed problems'') may be a more faithful parameterization. Second, the rubric was authored by us from data inspection and conversations with professional coders.

\begin{table*}[t]
\centering
\caption{The 10 rubric dimensions. S (for ICD-10-CM) and C (for CPT) dimensions are scored from (note, codes) pairs and can be applied at inference time.}
\label{tab:rubric}
%\resizebox{\textwidth}{!}{%
\begin{tabular}{lll}
\toprule
ID & Name & Scale ($1 \to 5$) \\
\midrule
S1 & Problem Scope & Chief complaint only $\to$ every documented problem \\
S2 & Inferential Aggressiveness & Only explicitly named $\to$ infer from labs/meds/context \\
S3 & Administrative Completeness & No Z-codes $\to$ all applicable Z-codes \\
S4 & Specificity Preference & Always unspecified $\to$ most specific inferrable \\
S5 & Active Management Threshold & Only if treatment changed $\to$ any documented condition \\
S6 & Comorbidity Inclusion & Only focal $\to$ all mentioned comorbidities \\
\midrule
C1 & Procedural Evidence Threshold & Full administrative details $\to$ inferred from context \\
C2 & E/M Level Philosophy & Lowest defensible $\to$ highest supportable \\
C3 & Modifier Application & Never unless required $\to$ apply liberally \\
C4 & ICD--CPT Linkage Scope & Single most relevant $\to$ all remotely related \\
\bottomrule
\end{tabular}
%}
\end{table*}

\begin{table*}[t]
\centering
\caption{Five data distributions (four outpatient; MIMIC-IV inpatient).}
\label{tab:data}
\resizebox{\textwidth}{!}{%
\begin{tabular}{lcll}
\toprule
Dist. & $N$ & Note & Coder \\
\midrule
CONV & 423 & Outpatient summarized conversations & Human coders, double-annotated + adjudicated, single-note context \\
EHR & 450 & Outpatient natural notes, multi-hospital & Mixed sources per encounter (human, possibly AI-assisted) \\
ACI-Y & 193 & Outpatient ACI-Bench encounters & Double-annotated + adjudicated ICD-10-CM \citep{yuan2025verify} \\
ACI-M & 193 & Outpatient ACI-Bench encounters & Single-coder ICD-10-CM from MedCodER \citep{baksi2024medcoder} \\
MIMIC & 450 & Inpatient MIMIC Discharge Summaries & Assigned from the full inpatient record \citep{johnson2023mimic}\\
\bottomrule
\end{tabular}
}
\end{table*}

\begin{table}[t]
\centering
\caption{Per-dataset style profiles (mean of 1--5 scores, std-dev in parentheses; ``--'' where C-dimensions are not exercised because CPT gold is absent). ACI-Y and ACI-M have nearly identical averaged profiles despite $0.73$ inter-annotator Jaccard similarity on the same notes; the main difference is S2 (inferential aggressiveness).}
\label{tab:dataset_profiles}
\setlength{\tabcolsep}{3pt}
\resizebox{\columnwidth}{!}{%
\begin{tabular}{l*{5}{Z}}
\toprule
Dimension & \multicolumn{1}{c}{CONV} & \multicolumn{1}{c}{EHR} & \multicolumn{1}{c}{ACI-Y} & \multicolumn{1}{c}{ACI-M} & MIMIC\\
\midrule
\multicolumn{6}{l}{ICD dimensions} \\
S1 Problem Scope       & 2.8 {\scriptsize(.9)} & 1.4 {\scriptsize(.7)} & 2.5 {\scriptsize(1.0)} & 2.2 {\scriptsize(1.0)} & 4.5 {\scriptsize(.8)} \\
S2 Inferential Aggr.   & 1.8 {\scriptsize(.8)} & 1.2 {\scriptsize(.5)} & 1.8 {\scriptsize(1.0)} & 1.4 {\scriptsize(.7)} & 3.5 {\scriptsize(.9)} \\
S3 Admin Completeness  & 2.4 {\scriptsize(1.0)} & 2.0 {\scriptsize(1.0)} & 2.0 {\scriptsize(1.1)} & 1.5 {\scriptsize(.9)} & 3.9 {\scriptsize(1.0)} \\
S4 Specificity Pref.   & 2.5 {\scriptsize(.7)} & 2.9 {\scriptsize(.8)} & 2.9 {\scriptsize(.8)} & 2.8 {\scriptsize(.9)} & 3.3 {\scriptsize(.6)} \\
S5 Active Mgmt Thresh. & 2.6 {\scriptsize(.7)} & 1.6 {\scriptsize(.8)} & 2.5 {\scriptsize(.8)} & 2.4 {\scriptsize(.9)} & 3.7 {\scriptsize(.6)} \\
S6 Comorbidity Incl.   & 2.1 {\scriptsize(.8)} & 1.0 {\scriptsize(.2)} & 1.8 {\scriptsize(.9)} & 1.9 {\scriptsize(.9)} & 4.0 {\scriptsize(.9)} \\
\midrule
\multicolumn{6}{l}{CPT dimensions} \\
C1 Procedural Evid.    & 1.7 {\scriptsize(.9)} & 2.5 {\scriptsize(.8)} & \multicolumn{1}{c}{--} &  \multicolumn{1}{c}{--} & \multicolumn{1}{c}{--} \\
C2 E/M Level Phil.     & 3.1 {\scriptsize(.3)} & 3.1 {\scriptsize(.6)} & \multicolumn{1}{c}{--} & \multicolumn{1}{c}{--} & \multicolumn{1}{c}{--} \\
C3 Modifier Appl.      & 1.0 {\scriptsize(.2)} & 1.1 {\scriptsize(.5)} & \multicolumn{1}{c}{--} & \multicolumn{1}{c}{--} & \multicolumn{1}{c}{--} \\
C4 ICD--CPT Linkage    & 2.7 {\scriptsize(.9)} & 1.2 {\scriptsize(.6)} & \multicolumn{1}{c}{--} & \multicolumn{1}{c}{--} & \multicolumn{1}{c}{--} \\
\bottomrule
\end{tabular}
}
\end{table}

\section{Setup}
\label{app:setup}

\subsection{Data}
\label{sec:data}

Five data distributions (CONV, EHR, ACI-Y, ACI-M, MIMIC-IV) of (note, code-set) pairs with ICD-10-CM codes (CPT codes for CONV and EHR only). The first four outpatient, MIMIC-IV \citep{johnson2023mimic} inpatient. They differ in patient care settings, note source, coder, and style $\psi_i$ (Table~\ref{tab:data}; per-dataset style profiles in Table~\ref{tab:dataset_profiles}). CONV and EHR are internal and anonymized; ACI-Y \citep{yuan2025verify} and ACI-M \citep{baksi2024medcoder} are built on the public ACI-Bench notes \citep{yim2023acibench}. Because ACI-Y and ACI-M share the same 110 notes, they isolate coder style from note content by construction.

\subsection{Style Methods}
\label{sec:methods}

For each dataset we hold out 150 encounters as a ``train'' split, score them with the profiler, and average the exercised dimensions into a single dataset-level style $\psi'$ used at test time. The base coder is a Zero-shot prompt (full text in Appendix~\ref{app:zeroshot}).

\begin{itemize}[noitemsep,topsep=2pt]
\item \textbf{Zero-shot (no style):} the Zero-shot coder with no style block; a lower reference point.
\item \textbf{Dataset Avg Style:} the averaged $\psi'$ derived from the target dataset's train split (the ``matched'' condition).
\item \textbf{Style Oracle:} $\psi'$ is replaced by a per-encounter profile scored from that encounter's own (note, gold codes); an upper bound on what style conditioning can recover when the gold coder's style is known exactly.
\item \textbf{Styles all-1s} and \textbf{Styles all-5s:} extreme controls that set every dimension to the minimum (1) or maximum (5) of the 1--5 scale, representing the conservative and aggressive ends of the coding spectrum. If style were descriptive noise, these would leave F1 unchanged.
\end{itemize}

\subsection{Conditioning the Model at Inference}
\label{app:prompt}

At inference the averaged profile $\psi'$ is rendered into a natural-language \texttt{<coding\_style>} block that is prepended to the user message. Each of the 10 dimensions emits one paragraph carrying its integer score (1--5) and a written behavior description for that score. For Problem Scope, 1 means ``code only the chief complaint'', 3 means ``chief complaint plus problems actively addressed today'', and 5 means ``every problem in the record''.

\begin{tcolorbox}[breakable, colback=gray!5, colframe=gray!50, boxrule=0.4pt, left=4pt, right=4pt, top=3pt, bottom=3pt]
{\footnotesize
\begin{verbatim}
<coding_style>
Apply the following preferences when deciding 
what and how to code this encounter.

=== ICD-10 CODING STYLE ===
PROBLEM SCOPE (3/5, if applicable):
Code the chief complaint plus all problems that 
were actively addressed today (a clinical action 
was taken or decision was made).

[...continues for all 10 dimensions...]
</coding_style>
\end{verbatim}
}
\end{tcolorbox}

\section{Experiments}
\label{sec:experiments}

All reported F1 numbers are sample-averaged: we compute a set-based F1 between predicted and gold ICD codes per encounter and take the unweighted mean over encounters. The overall F1 is the mean over encounters pooled across datasets (i.e., doc-weighted, not the mean of per-dataset means). Anthropic Opus 4.8 is used for all prompt based methods.

\subsection{Separability: Style Separates Data}
\label{sec:separability}
\begin{figure*}[t]
\centering
\includegraphics[width=0.7\textwidth]{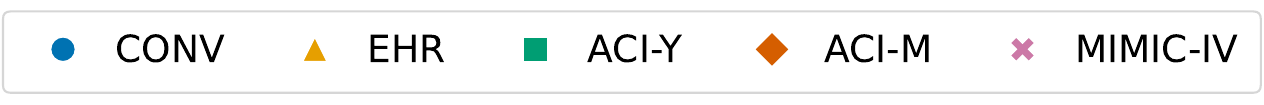}\\[-4pt]
\subfigure[ICD only (S1--S6)]{%
  \includegraphics[width=0.48\textwidth]{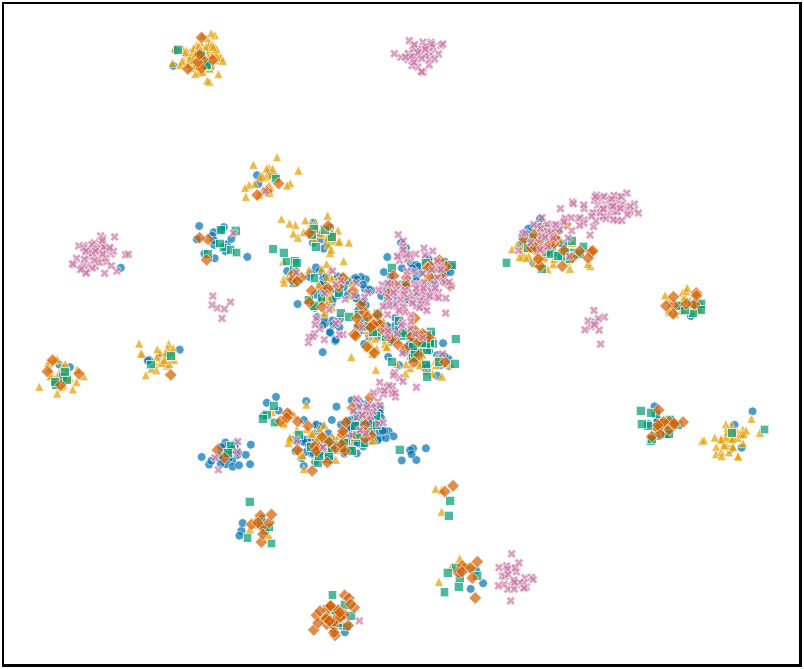}%
  \label{fig:umap-S}%
}\hfill
\subfigure[CPT only (C1--C4), CONV/EHR only]{%
  \includegraphics[width=0.48\textwidth]{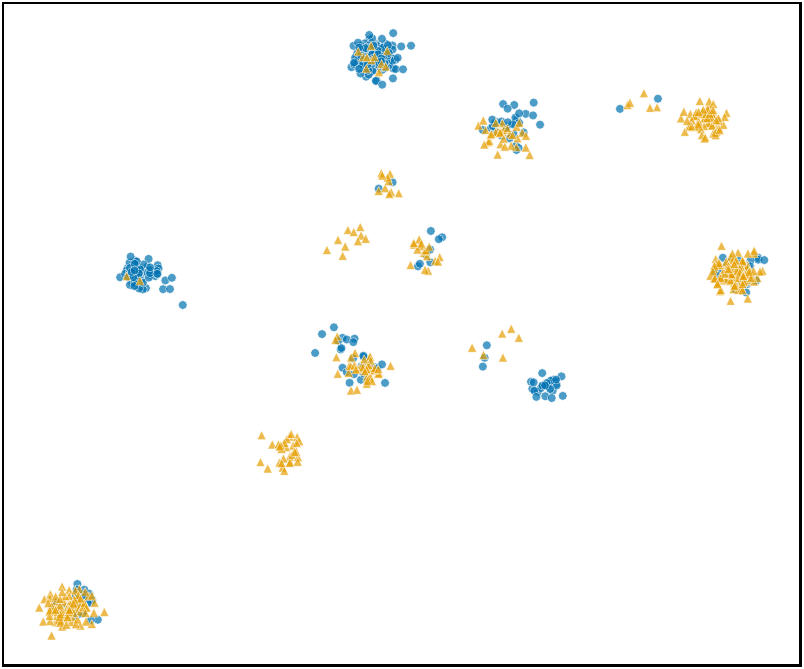}%
  \label{fig:umap-C}%
}
\caption{UMAP of per-encounter style vectors, colored and shaped by dataset. (a) ICD dimensions alone (S1--S6), including MIMIC-IV: datasets show visible regional biases but substantial mixing. (b) CPT dimensions alone (C1--C4), restricted to CONV/EHR which have CPT codes.}
\label{fig:tsne}
\end{figure*}

If $\psi$ is a real property of a data source rather than per-note noise, per-encounter style vectors should cluster by dataset. We project the six ICD-only style dimensions and the four CPT-only style dimensions in isolation with UMAP \citep{mcinnes2018umap} (Figure~\ref{fig:tsne}), so that ICD-coding style and CPT-coding style are evaluated as separate signals.

\subsection{Steerability: ICD Conditioning on Style}
\label{sec:controllability}

We condition the coder on a style profile estimated using various strategies, plus two extreme profiles (all-1s and all-5s, representing the conservative and aggressive ends of the coding spectrum) as a control: if style were merely descriptive noise, injecting an extreme profile would leave F1 unchanged (Table~\ref{tab:controllability}).

\begin{table}[h!]
\centering
\caption{Per-dataset ICD F1 by conditioning strategy. Matched style helps every outpatient dataset (most on the narrow-scope EHR, $+26.4$) but not inpatient MIMIC-IV ($-2.7$; \S\ref{sec:controllability}), per-sample oracle profiles help further, and the extreme all-5s style drops F1 by 10--21 points everywhere, showing the model follows the injected style.}
\label{tab:controllability}
\setlength{\tabcolsep}{4pt}
\resizebox{\columnwidth}{!}{%
\begin{tabular}{l*{5}{F}}
\toprule
Condition & \multicolumn{1}{c}{CONV} & \multicolumn{1}{c}{EHR} & \multicolumn{1}{c}{ACI-Y} & \multicolumn{1}{c}{ACI-M} & MIMIC \\
\midrule
Zero-shot (no style) & 44.5 & 28.8 & 47.7 & 45.8 & 48.8 \\
Dataset Avg Style & 59.1 & 55.2 & 69.6 & 65.9 & 46.1 \\
Styles all-1s & 35.2 & 51.1 & 50.7 & 54.4 & 10.7 \\
Styles all-5s & 31.0 & 14.4 & 27.5 & 25.3 & 38.4 \\
\midrule
Style Oracle & 62.8 & 58.0 & 70.7 & 70.9 & 47.9 \\
\bottomrule
\end{tabular}
}
\end{table}

The extreme styles move F1 sharply: all-5s reduces every outpatient dataset by 14--21 points, and all-1s costs $-9$ on CONV while helping the narrow-scope EHR ($+22$), as we would expect from a minimal-coding profile. Matching the dataset style raises ICD F1 on every outpatient corpus (up to $+26.4$ on EHR) and per-sample oracle profiles (an upper bound using the style profiled on the ground truth codes) add more (up to $+29$ on EHR). Inpatient MIMIC-IV is the exception, and we discuss it separately below.

\textbf{When matched style does not help: inpatient MIMIC-IV.} On inpatient MIMIC-IV, matched style does not improve ICD F1 ($48.8\to46.1$; oracle $47.9$), yet the mechanism still operates: the extreme controls move F1 sharply (all-1s $10.7$, all-5s $38.4$), suggesting that it is the \emph{matched} profile, not the coder, that fails to align. We suspect a rubric-range mismatch. Our rubric was derived from outpatient data ($2.3$ codes per encounter), so applying it to MIMIC-IV ($14.5$ codes per encounter; S1 $4.5$, S6 $4.0$; Table~\ref{tab:dataset_profiles}) leaves little headroom to steer further. See Section~\ref{sec:discussion} for further discussion.

\subsection{Steerability: CPT Conditioning on Style}
\label{sec:cpt}

CPT tells a different story (Table~\ref{tab:cpt}; ACI-Y/ACI-M lack CPT gold annotations).
Matching a dataset's style improves F1 on CONV by $+5.4$ points but leave EHR essentially unchanged ($-1.6$). This is in contrast to the large ICD gains of up to $+24$ (Table~\ref{tab:controllability}).
Extreme profiles still cause large drops,
so the coder does follow CPT instructions when they are informative.
We hypothesize there is simply less CPT style variation to exploit:
two of the four CPT dimensions have nearly identical means across CONV/EHR
(Table~\ref{tab:dataset_profiles},
and while C1 and C4 do differ, the matched profile averages these out.
Matched profiles can only help to the extent that profiles differ across corpora. This suggests that in these corpora style variation lives mostly in ICD scope decisions rather than in procedure coding or that the style dimensions and/or model's ability to apply them works less well for CPT.

\begin{table}[h!]
\centering
\caption{Per-dataset CPT F1 by conditioning strategy (Zero-shot coder). Matched and oracle styles move CPT F1 by at most $+10$ points on CONV and $+1.3$ on EHR. Extreme profiles cause dramatic drops, especially all-5s on CONV ($-54.6$), demonstrating that the coder follows CPT style instructions when they are informative.}
\label{tab:cpt}
\setlength{\tabcolsep}{4pt}
\begin{tabular}{l*{2}{F}}
\toprule
Condition & \multicolumn{1}{c}{CONV} & \multicolumn{1}{c}{EHR} \\
\midrule
Zero-shot (no style) & 60.2 & 59.9 \\
Dataset Avg Style & 65.6 & 58.3 \\
Styles all-1s & 23.2 & 51.9 \\
Styles all-5s & 5.6 & 35.2 \\
\midrule
Style Oracle & 70.1 & 61.2 \\
\bottomrule
\end{tabular}
\end{table}

\begin{table*}[t]
\centering
\caption{Per-dataset ICD F1 for Zero-shot and Optimized ZS, Two-Step \citep{boyle2023automated} and Yuan-Verify \citep{yuan2025verify}, plus MIMIC-IV ICD-10 supervised baselines CAML and PLM-ICD from \citet{edin2023mimic}. Matched style improves outpatient dataset but is neutral on inpatient MIMIC-IV (\S\ref{sec:controllability}); the extreme all-5s profile collapses F1, showing the models follow the injected style. CAML and PLM-ICD have no style input, so only a single column is reported for each. The Average row is a micro-average (per-encounter F1 pooled across the 5 test sets, $n=1{,}093$). $^\star$ marks significance in the Average row by an unpaired Welch $t$-test comparing matched vs.\ no-style per-encounter F1 ($p<0.05$; Appendix~\ref{app:ttest}).}
\label{tab:methods_perdataset}
\setlength{\tabcolsep}{3pt}
%\resizebox{\textwidth}{!}{%
\begin{tabular}{l*{2}{F}@{\hspace{1.2em}}*{2}{F}@{\hspace{1.2em}}*{2}{F}@{\hspace{1.2em}}*{2}{F}@{\hspace{1.2em}}FF}
\toprule
& \multicolumn{2}{c}{Zero-shot} & \multicolumn{2}{c}{Optimized ZS} & \multicolumn{2}{c}{Two-Step} & \multicolumn{2}{c}{Yuan-Verify} & \multicolumn{1}{c}{CAML} & \multicolumn{1}{c}{PLM-ICD} \\
\cmidrule(lr){2-3}\cmidrule(lr){4-5}\cmidrule(lr){6-7}\cmidrule(lr){8-9}\cmidrule(lr){10-10}\cmidrule(lr){11-11}
Dataset & \multicolumn{1}{c}{no style} & \multicolumn{1}{c}{matched} & \multicolumn{1}{c}{no style} & \multicolumn{1}{c}{matched} & \multicolumn{1}{c}{no style} & \multicolumn{1}{c}{matched} & \multicolumn{1}{c}{no style} & \multicolumn{1}{c}{matched} & \multicolumn{2}{c}{Supervised (no style)} \\
\midrule
CONV  & 44.5 & 59.1 & 57.0 & 59.8 & 45.1 & 55.2 & 45.0 & 57.4 & 18.1 & 28.3 \\
EHR   & 28.8 & 55.2 & 38.5 & 56.1 & 24.2 & 51.2 & 30.8 & 49.9 &  5.0 & 11.4 \\
ACI-Y & 47.7 & 69.6 & 60.9 & 65.7 & 49.5 & 65.9 & 49.6 & 65.1 & 12.6 & 29.6 \\
ACI-M & 45.8 & 65.9 & 60.1 & 64.1 & 48.1 & 63.6 & 48.1 & 65.5 & 12.0 & 29.4 \\
MIMIC & 48.8 & 46.1 & 45.5 & 46.5 & 41.1 & 41.1 & 47.3 & 45.7 & 48.9 & 50.1 \\ 
\midrule
Average & 41.8 & 56.2$^\star$ & 49.5 & 56.2$^\star$ & 39.0 & 52.1$^\star$ & 42.5 & 53.7$^\star$ & 21.8 & 29.9 \\
\bottomrule
\end{tabular}
%}
\end{table*}

\subsection{Same-Note Cross-Annotator (ACI-Y/ACI-M)}
\label{sec:crossannotator}

ACI-Y and ACI-M annotate the same 110 notes independently, and their inter-annotator Jaccard similarity is only $73\%$. To isolate style from coding error, we had every ICD-10 code in both annotations independently audited by clinical coders and marked defensible or not (\S\ref{app:audit}); after erroneous codes are removed, the two annotators still agree on only $77\%$ of their codes. Because the remaining codes are individually defensible on the note, this residual gap is style, not clinical disagreement. Yet the averaged rubric profiles we recover for ACI-Y and ACI-M are nearly identical (Table~\ref{tab:dataset_profiles}):
 they agree on five of six ICD dimensions and differ mainly on S2 ($1.76$ vs.\ $1.40$). We take this as a rubric-coverage limitation: the ten dimensions were designed on CONV/EHR and do not fully cover the axes on which these two annotator groups disagree. Nevertheless, per-sample oracle profiles still recover $+23$ points against either gold (Table~\ref{tab:crossannotator}), all-1s styles are roughly neutral ($+3$ vs.\ ACI-Y, $+9$ vs.\ ACI-M), and opposite (all-5s) styles drop F1 by $-20$: style operates at the per-note decision level even when aggregate profiles cannot distinguish the annotators.

\begin{table}[h!]
\centering
\caption{ICD F1 on the shared ACI-Bench notes across different annotators. Per-sample profiles recover $+23$ points against either gold despite nearly identical aggregate styles: style acts at the per-note level.}
\label{tab:crossannotator}
\resizebox{\columnwidth}{!}{%
\begin{tabular}{l*{2}{F}}
\toprule
Condition & \multicolumn{1}{c}{vs.\ ACI-Y gold} & \multicolumn{1}{c}{vs.\ ACI-M gold} \\ 
\midrule
Zero-shot (no style) & 47.7 & 45.8 \\
Dataset Avg Style& 67.5 & 67.5 \\
Styles all-1s & 50.7 & 54.4 \\
Styles all-5s & 27.5 & 25.3 \\
\midrule
Style Oracle & 70.7 & 68.5 \\
\bottomrule
\end{tabular}
}
\end{table}

\subsection{Generalizability: Augmenting Existing Methods}
\label{sec:generalizability}

If style is a property of the data rather than of our coder, style conditioning should help any coding method. We test this against two published prompt-based LLM coding pipelines as baselines: the two-step pipeline of \citet{boyle2023automated} and the verify pipeline of \citet{yuan2025verify}, running each with and without the injected style block. As an additional reference point, we also include Optimized ZS: our best attempt at prompt-engineering a zero-shot coder without style conditioning, representing a ceiling of what prompt design alone can achieve. As supervised references, we also report two MIMIC-IV ICD-10 baselines from the replicability study of \citet{edin2023mimic}: CAML \citep{mullenbach2018explainable} and PLM-ICD \citep{huang2022plmicd}, which take no style input and appear only in the no-style condition.
In Table~\ref{tab:methods_perdataset} we observe matched style improves every method on the outpatient corpora (it is neutral on inpatient MIMIC-IV; \S\ref{sec:controllability}), and the minimal-prompt Zero-shot coder gains the most, landing at $60.0$ F1, indistinguishable from Optimized ZS+matched ($59.8$). Prompt engineering (Zero-shot $\rightarrow$ Optimized ZS) achieves $+11.8$ F1 by itself; style conditioning alone matches or exceeds that gain and is orthogonal to the coding method.

\begin{figure}[t]
\centering
\includegraphics[width=\columnwidth]{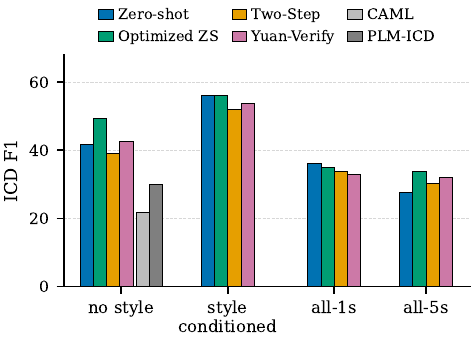}
\caption{Style dominates ICD F1 across four style conditions. Within each style group, the four LLM coders (colored bars) sit at nearly the same height: the injected style, not the choice of coding method, dominates F1. Style conditioning lifts every method to $\sim$57--60 F1. CAML and PLM-ICD (supervised MIMIC references, no style input) are shown only in the no-style group.}
\label{fig:generalizability}
\end{figure}

The extreme mismatched styles serve as the control here: an all-5s profile collapses F1 for every method, so all four follow the injected style rather than ignoring it. The all-1s profile hurts Optimized ZS but helps the other methods. Coders that were not heavily prompt-optimized are more receptive to any style instruction.

\section{Discussion}
\label{sec:discussion}

\textbf{Style has real impact.} Separable clusters, matched-style gains, and large wrong-style drops all support treating coding style as a latent variable rather than annotation noise; the strongest single piece of evidence is that ACI-Y and ACI-M share only $73\%$ of their codes (Jaccard) on identical notes, and that an independent clinical audit lifts this only to $77\%$ after erroneous codes are removed. We do not claim the ten rubric dimensions are the one true set of styles, only that they capture enough of the variation to move F1 substantially.

\textbf{The model has sufficient coding knowledge; style unlocks more.} Once an appropriate profile is supplied, our zero-shot coder, Optimized ZS, Two-Step, and Yuan-Verify converge to within 4 F1 points. Strikingly, Zero-shot+matched ($60.0$) matches Optimized ZS+matched ($59.8$): style conditioning can be viewed as a controlled form of prompt optimization.

\textbf{A lower bound.} The rubric closes a substantial fraction of the F1 gap but does not capture all of style: ACI-Y and ACI-M share only $73\%$ of their codes ($77\%$ after an independent clinical audit removes errors) yet have nearly identical averaged profiles (Table~\ref{tab:dataset_profiles}), so additional style axes exist beyond the current ten. Inpatient MIMIC-IV is a second, independent witness: its profile saturates the top of the outpatient-derived $1$--$5$ scale, and matched style does not help there even though the extreme styles still move F1 (\S\ref{sec:controllability}).

\textbf{Evaluation.} Single-gold F1 underestimates LLM capability on style-mismatched data. Reporting style-conditioned F1, or at least characterizing the corpus $\psi'$, would make coding evaluations more comparable. %\han{not a comment, but I really like this}

\textbf{Style vs.\ compliance.} We use ``style'' descriptively. Some preferences sit within the guidelines (Z-code completeness, specificity); others approach the boundary of compliance (e.g., coding a diagnosis documented only as probable). Our framework measures these preferences; it does not endorse them.

\bibliography{ref}

% \newpage
% .
% \newpage
\appendix

\section{ACI-Bench code defensibility audit}
\label{app:audit}

To isolate coding style from coding error, we conducted an independent human audit of the ICD-10 codes on the shared ACI-Bench notes. Two clinical coders reviewed every ICD-10 code produced by ACI-Y and ACI-M side-by-side per note, marking each individual code as \emph{defensible} or \emph{not defensible}; a free-text reason was required for every not-defensible verdict. 
Across the 110 notes audited, ACI-Y's codes are $89.8\%$ defensible ($202/225$) and ACI-M's are $86.4\%$ ($197/228$): the vast majority of each annotator's output is a supportable coding choice on the note. Recomputing the vendor-vs-vendor overlap on the audit-verified subset (i.e., removing not-defensible codes from each side before taking the set intersection and union) raises the per-note Jaccard from $73\%$ to $77\%$.

\section{Significance test for Table~\ref{tab:methods_perdataset}}
\label{app:ttest}

For each of the four LLM coders we compare matched-style and no-style per-encounter ICD F1 with an unpaired two-sided Welch $t$-test (\texttt{scipy.stats.ttest\_ind}, \texttt{equal\_var=False}) on the pooled test-set encounters: CONV $273$, EHR $300$, ACI-Y $110$, ACI-M $110$, MIMIC $300$, total $n=1{,}093$ per condition. CAML and PLM-ICD have no matched variant and are not tested.

\begin{table}[h!]
\centering
\caption{Unpaired Welch $t$-test of matched-style vs no-style per-encounter ICD F1 pooled over the 5 test sets ($n=1{,}093$ per condition).}
\label{tab:ttest}
\resizebox{\columnwidth}{!}{%
\begin{tabular}{lrrrrr}
\toprule
Coder & F1$_\text{ns}$ & F1$_\text{m}$ & $\Delta$ & $t$ & $p$ \\
\midrule
Zero-shot     & $41.83$ & $56.20$ & $+14.37$ & $11.40$ & $3.1\times 10^{-29}$ \\
Optimized ZS  & $49.47$ & $56.18$ &  $+6.71$ &  $4.97$ & $7.1\times 10^{-7}$   \\
Two-Step      & $39.02$ & $52.14$ & $+13.12$ & $10.12$ & $1.6\times 10^{-23}$  \\
Yuan-Verify   & $42.53$ & $53.70$ & $+11.17$ &  $8.48$ & $4.1\times 10^{-17}$  \\
\bottomrule
\end{tabular}
}
\end{table}

\section{Materials}
\label{sec:materials}

\subsection{Zero-shot Coding Prompt}
\label{app:zeroshot}

The Zero-shot (no-style) coder uses a single fixed system prompt and user template, with no exemplars, no chain-of-thought scaffolding, and no post-processing. It is the same prompt used throughout the paper as the ``Zero-shot'' / ``no style'' condition; the style block described in Appendix~\ref{app:prompt} is prepended to the user message when a style is applied.

\paragraph{Prompt.} The placeholder \texttt{\{clinical\_note\}} is replaced with the raw note text; no other formatting is applied.

\begin{tcolorbox}[breakable, colback=gray!5, colframe=gray!50, boxrule=0.4pt, left=4pt, right=4pt, top=3pt, bottom=3pt]
{\footnotesize
\begin{verbatim}
<task>
Generate billable medical codes
(ICD-10-CM, CPT) from the clinical note.
</task>

<output_format>
{
  "icd-10-cm": [
    {"code": "ICD-10-CM code",
     "description": "description",
     "type": "primary|secondary"}
  ],
  "cpt": [
    {"code": "CPT code",
     "description": "description",
     "type": "E/M|Other"}
  ]
}
</output_format>

<clinical_note>
{clinical_note}
</clinical_note>
\end{verbatim}
}
\end{tcolorbox}

\section{Qualitative Examples}
\label{app:qualitative}

\begin{figure*}[h!]
\centering
\includegraphics[width=\textwidth]{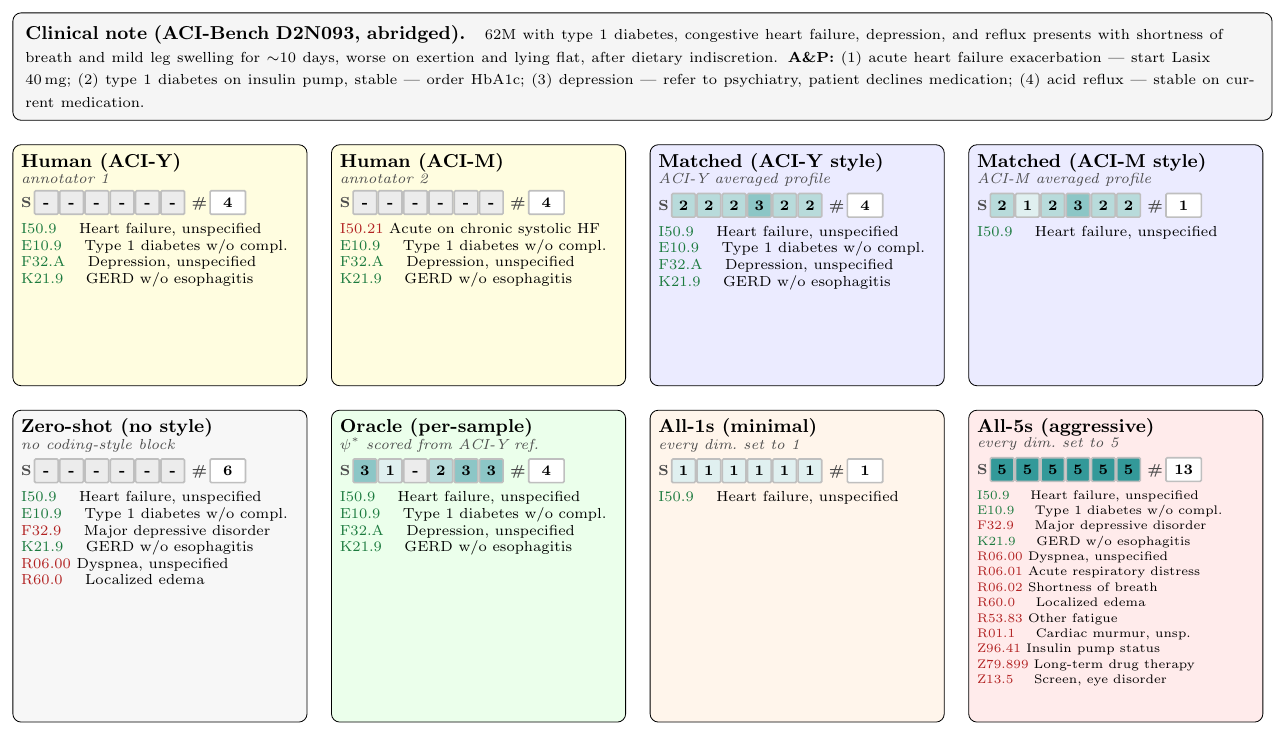}
\caption{An ACI-Bench encounter (D2N093) coded by two human annotators (ACI-Y, ACI-M) and by the model under five style conditions (Zero-shot (no-style), Matched to ACI-Y, Matched to ACI-M, Oracle, All-1s, All-5s). The top box shows an abridged version of the clinical note. Each panel's header shows the six ICD-scope style scores S1--S6 (pale = 1, saturated = 5, gray = not applicable) injected into the prompt and the total number of codes~(\#). Green marks codes that appear in the ACI-Y annotation; red marks codes that do not. Aggregate results are in Table~\ref{tab:controllability} (main paper) and Table~\ref{tab:methods_perdataset} (appendix).}
\label{fig:conditions_example}
\end{figure*}

\begin{figure*}[h!]
\centering
\includegraphics[width=\textwidth]{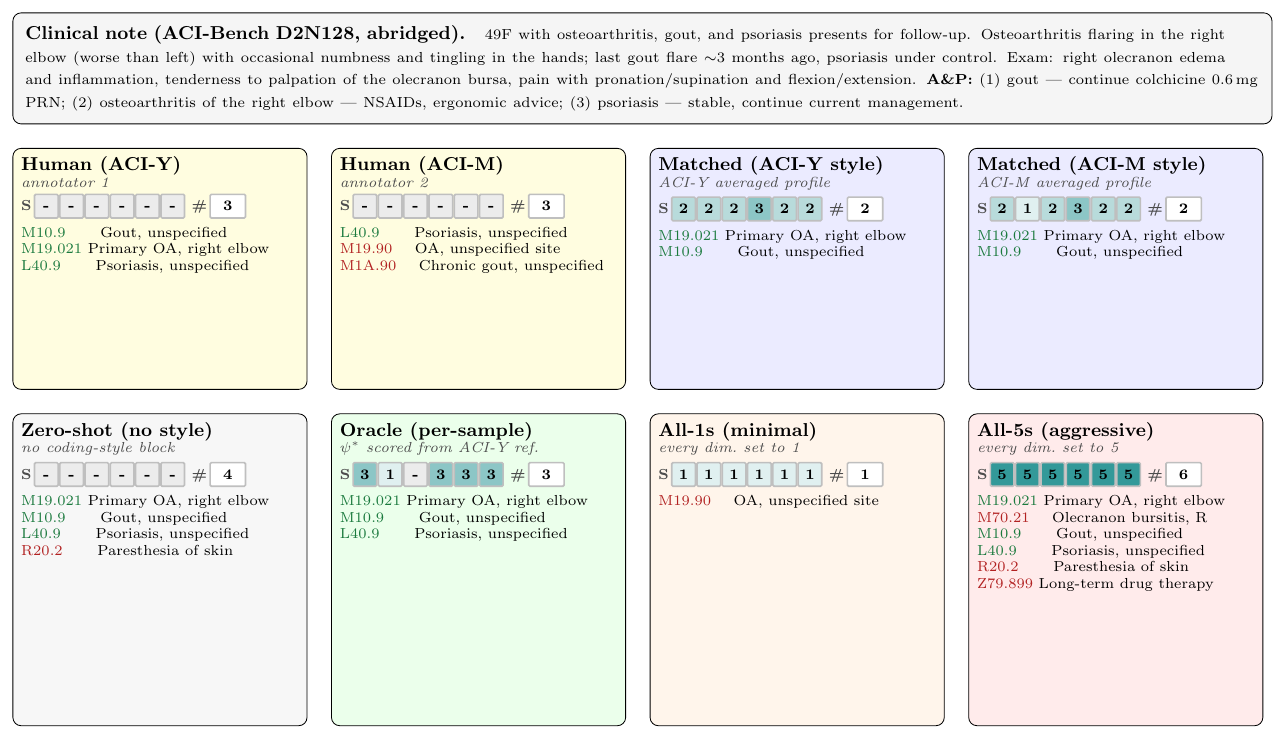}
\caption{A second ACI-Bench encounter (D2N128), same layout as Figure~\ref{fig:conditions_example}: the abridged note at top, then two human annotations (ACI-Y, ACI-M) and the model's outputs under five style conditions (Baseline, Matched to ACI-Y, Matched to ACI-M, Oracle, All-1s, All-5s). Each panel's header shows the six ICD-scope style scores S1--S6 (pale = 1, saturated = 5, gray = not applicable) and the total number of codes~(\#). Green marks codes that appear in the ACI-Y annotation; red marks codes that do not.}
\label{fig:conditions_example_ex2}
\end{figure*}

Figures~\ref{fig:conditions_example} and~\ref{fig:conditions_example_ex2} show two ACI-Bench encounters coded under all five style conditions and by both human annotators.

\end{document}